# EHRNavigator: A Multi-Agent System for Patient-Level Clinical Question Answering over Heterogeneous Electronic Health Records

Lingfei Qian*, Mauro Giuffre*, Yan Wang, Huan He, Qianqian Xie, Xuguang Ai, Xueqing Peng, Fan Ma, Ruey-Ling Weng, Donald Wright, Adan Wang, Qingyu Chen, Vipina K. Keloth, Hua Xu*

**Abstract**

Clinical decision-making increasingly relies on timely and context-aware access to patient information within Electronic Health Records (EHRs), yet most existing natural language question-answering (QA) systems are evaluated solely on benchmark datasets, limiting their practical relevance. To overcome this limitation, we introduce EHRNavigator, a multi-agent framework that harnesses AI agents to perform patient-level question answering across heterogeneous and multimodal EHR data. We assessed its performance using both public benchmark and institutional datasets under realistic hospital conditions characterized by diverse schemas, temporal reasoning demands, and multimodal evidence integration. Through quantitative evaluation and clinician-validated chart review, EHRNavigator demonstrated strong generalization, achieving 86% accuracy on real-world cases while maintaining clinically acceptable response times. Overall, these findings confirm that EHRNavigator effectively bridges the gap between benchmark evaluation and clinical deployment, offering a robust, adaptive, and efficient solution for real-world EHR question answering.

## 1. Introduction

Electronic Health Records (EHRs) serve as comprehensive repositories of patient medical history, encompassing structured data such as laboratory results and medication orders, unstructured data such as free-text clinical notes, and medical images [31]. In routine clinical practice, clinicians frequently need to query EHRs to extract relevant, time-sensitive information for decision-making [1]. For example, monitoring patients on broad-spectrum antibiotics requires synthesis of subjective clinical assessments, vital signs, and temporal trajectories in laboratory results. However, the volume, fragmentation, and heterogeneity of EHR data make such tasks time-consuming and error-prone, often leading to incomplete information retrieval, particularly when temporal reasoning or multi-source integration is required. Studies show that physicians spend an average of 16 minutes and 14 seconds per encounter using EHRs, with one-third of that time devoted to chart review alone [32]. A scoping review highlighted that technical issues (e.g., poor interoperability, data integrity challenges) and usability problems (e.g., fragmented displays, inefficient inbox management) impair clinicians' ability to track and interpret key patient information across the diagnostic workflow [27]. These challenges reflect the misalignment between EHR system design and clinicians' cognitive needs, underscoring the importance of more efficient and intelligent query mechanisms.

At the same time, the sheer amount of patient-level data contained in modern EHR systems further complicates manual review. For instance, patients in MIMIC-III, an intensive care EHR database sourced from Beth Israel Deaconess Medical Center, have an average of 45.1 clinical notes totaling more than 20,000 words, with some records exceeding 149,000 words (nearly 300 pages) [2,3]. Beyond unstructured notes, each patient also has extensive structured data spanning medications, laboratory results, procedures, and other clinical events. Manually reviewing such large and heterogeneous datasets is inefficient and unsustainable, reinforcing the need for automated systems that can rapidly surface relevant information for clinical question answering.

To support efficient natural language question answering (QA) over EHRs, researchers have explored a broad spectrum of natural language processing (NLP) approaches, ranging from traditional QA models to recent large language models (LLM) driven systems [5,6]. Early efforts primarily focused on patient-level questions over structured data, with systems such as TREQA [17], MAC-SQL [35], and EHR-SeqSQL [9] aiming to improve SQL generation accuracy through large-scale training on question–SQL pairs. More recent work, including EHRAgent [4] and EHR-RAG [13], sought to reduce manual schema engineering by enabling models to infer database structure with minimal human input, thereby improving generalization across diverse EHR schemas. In parallel, a separate line of research targets unstructured clinical notes using retrieval-based methods centered largely on keyword or embedding-based semantic matching [28,29,37]. However, most of these existing systems typically operate in isolation, retrieving relevant snippets without a unified reasoning framework to align them with structured clinical events. This often leads to a modality gap where unstructured notes are queried without the temporal or contextual guidance of relevant medical events provided by structured tables [45, 46]. Although recent research has extended EHR-based QA to multimodal settings by integrating structured data with clinical text [16], the proposed methods still rely on predefined classification rules to select which information to use, which is impractical in real-world clinical workflows.

Despite these efforts, significant limitations still hinder the translation of current research prototypes into real clinical workflows. **First**, most existing systems rely heavily on predefined question templates or schema-specific fine-tuning [4,6,9,10,11], which constrains flexibility and limits generalization across diverse institutional databases. Even when standardized schemas are used, differences in documentation practices across sites introduce additional barriers to cross-institutional adaptability [44]. **Second**, many approaches operate on only a single data modality, either structured tables or unstructured notes, thereby missing the complementary strengths inherent in multimodal EHR evidence. **Third**, widely used benchmark datasets predominantly emphasize point-in-time queries (e.g., "*What was the patient's WBC count on admission?*") while overlooking temporally grounded questions crucial for clinical reasoning (e.g., "*How did the WBC trajectory change during the first week of therapy?*"). **Finally**, most evaluation practices remain narrowly focused on reporting accuracy on benchmark datasets such as MIMIC, with virtually no assessment of practical deployment factors. Notably, existing studies rarely examine system latency issues that critically determine whether these systems can be used at the bedside. Consequently, current EHR question-answering systems struggle to generalize across institutions, perform temporal reasoning, and operate effectively under the performance constraints of live clinical settings.

To bridge these gaps, we present EHRNavigator, an end-to-end multi-agent framework designed for real-time, patient-level QA across heterogeneous EHR data. By orchestrating a coordinated network of specialized agents and external tools, EHRNavigator facilitates autonomous reasoning across both structured and unstructured modalities. This agentic architecture allows the system to decompose the complex text-to-SQL process into modular sub-tasks, invoke specialized tools on demand, and synthesize outputs, mirroring the diagnostic workflow of a clinician. As illustrated in Figure 1(a), EHRNavigator functions as an evidence-centric engine that systematically synthesizes findings from disparate tables and narratives. Rather than providing a simple answer, the framework generates evidence-backed responses paired with transparent reasoning pathways, allowing clinicians to trace every insight back to its source. Furthermore, EHRNavigator operates across diverse database structures without requiring schema-specific training, ensuring robust interoperability. Importantly, this work represents the first evaluation of an EHR-QA system to assess longitudinal trajectory reasoning on real-world clinical data. By rigorously analyzing implementation factors, including response time and system latency in a live hospital environment, EHRNavigator establishes a practical blueprint for the deployment of autonomous AI agents within high-stakes clinical workflows.

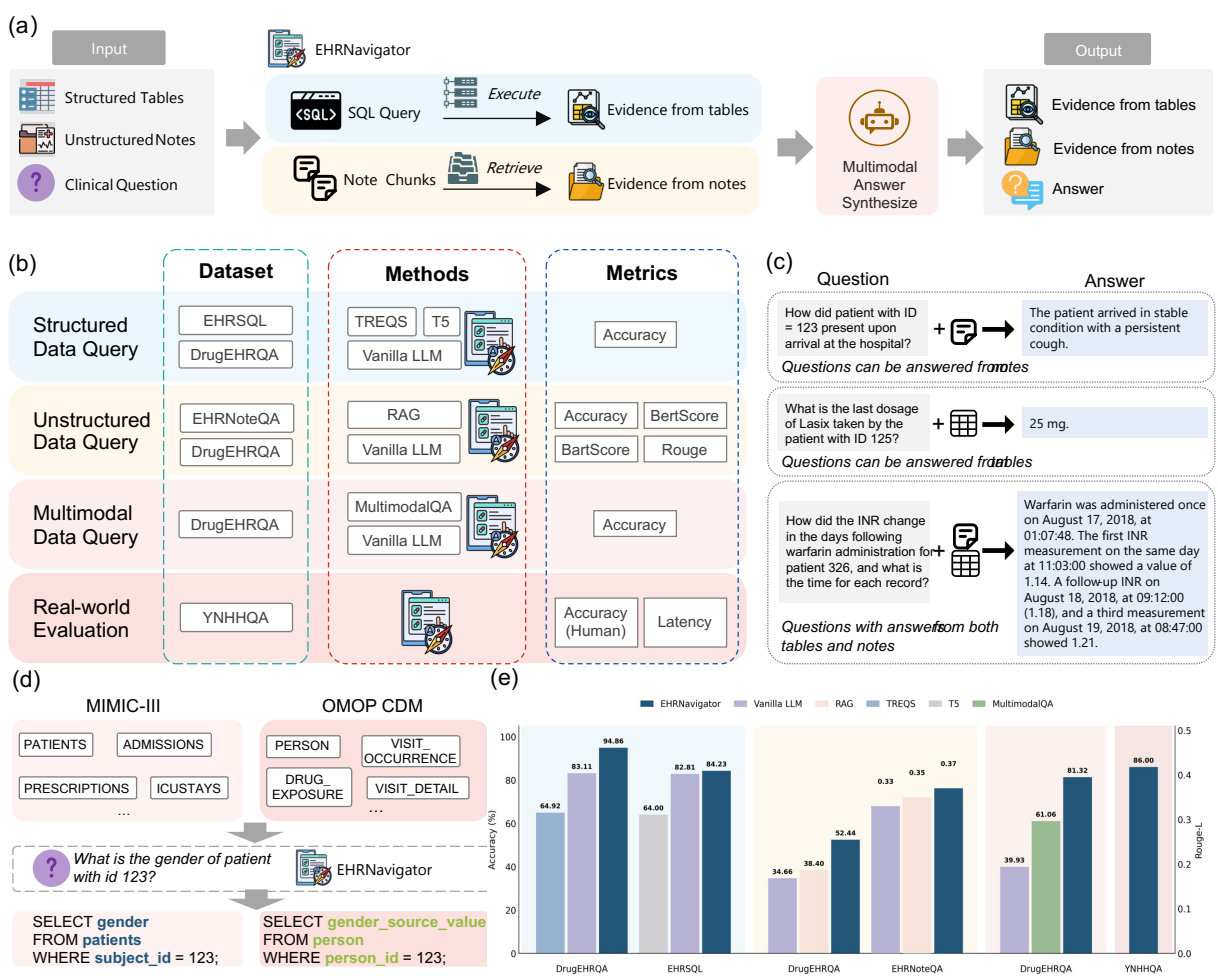

**Figure 1**: Overview of the EHRNavigator framework and evaluation protocol. (a) EHRNavigator System Architecture. Multi-agent orchestration designed for context-aware processing of structured, unstructured, and multimodal clinical queries. (b) Task Matrix and Modeling Paradigms. Systematic evaluation across four datasets (EHRSQL, EHRNoteQA, DrugEHRQA, and YNHHQA). (c) Clinical QA Modalities. Illustrative instances of evidence retrieval and synthesis derived from structured tables, clinical narratives, or cross-modal integration. (d) Cross-Schema Adaptation. Demonstration of EHRNavigator's capability to interpret heterogeneous architectures (e.g., MIMIC-III vs. OMOP CDM). The system dynamically maps clinical concepts to institution-specific schemas without manual retraining. (e) Performance Benchmarks. Comparative results validating that EHRNavigator significantly outperforms vanilla LLMs and RAG baselines in both execution accuracy and generative semantic alignment across (ROUGE-L for EHRNoteQA) across all benchmarks.

## 2. Results

This section presents the evaluation results of EHRNavigator, beginning with a system overview and the experimental setup, followed by a detailed analysis of its performance across multiple benchmarks.

### 2.1 System architecture and evaluation protocol

The comprehensive evaluation protocol is summarized in Figure 1(b) and (c). Our assessment spans four complementary datasets, EHRSQL, EHRNoteQA, DrugEHRQA, and YNHHQA (Yale New Haven Hospital question answering, which are curated for longitudinal trajectory reasoning), covering structured, unstructured, and multimodal clinical information. This setup evaluates three core query types in a real-world setting, characterizing model behavior across SQL-based retrieval, free-text reasoning, and cross-modal grounding. We benchmark EHRNavigator against four categories of methods: vanilla LLMs, RAG systems, pretrained language models (PLMs), and traditional text-to-SQL baselines.

Furthermore, Figure 1(d) shows EHRNavigator's cross-schema generalizability. Clinical databases often exhibit significant architectural divergence; for instance, the same clinical query may require entirely different table joins and column mapping in MIMIC-III versus OMOP CDM. EHRNavigator demonstrates robust adaptation to these heterogeneous schemas without institution-specific fine-tuning. Finally, the aggregate results in Figure 1(e) show that EHRNavigator significantly exceeds baseline performance across all four evaluation settings, validating its effectiveness under realistic operational constraints.

### 2.2 EHRNavigator boost multimodal clinical QA accuracy

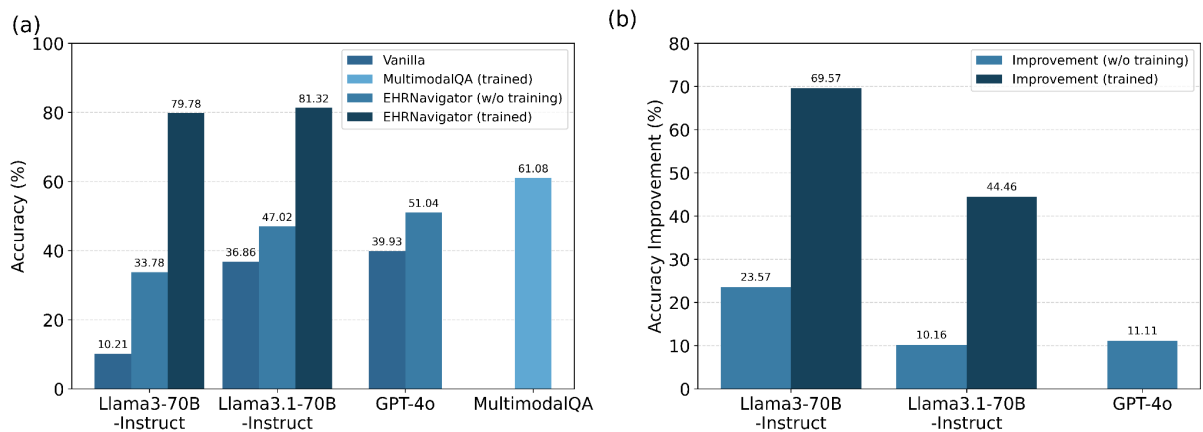

**Figure 2.** Comparative analysis of LLM performance in vanilla setting versus the EHRNavigator multi-agent decomposition framework. (a): Accuracy results for the vanilla setting, the EHRNavigator, and EHRNavigator (trained) configurations, juxtaposed with the MultimodalQA Baseline (a fine-tuned model utilizing ClinicalBERT and trained on DrugEHRQA). (b): Relative improvement in accuracy demonstrated by the EHRNavigator system over direct vanilla prompting.

We first evaluate EHRNavigator's performance on complex clinical queries that necessitate the seamless integration of structured records, and unstructured narratives from clinical notes. In real-world practice, these modalities often exist in silos, creating a "modality gap" that traditional models struggle to bridge.

To establish a robust comparison, we benchmark EHRNavigator against MultimodalQA [16], a specialized pre-trained language model using ClinicalBERT [47] as its backbone, designed specifically for cross-modal clinical evidence synthesis. In addition, we further examine three state-of-the-art LLMs, including GPT-4o [24], LLaMA3-70B-Instruct [25], and LLaMA3.1-70B-Instruct [25], under three progressively refined configurations. First, in the Vanilla setting, models are provided with the entire discharge summary and structured table data as raw context, relying solely on their inherent reasoning without agent-driven decomposition. This is contrasted with the EHRNavigator (no training), where models are integrated into our multi-agent layer to autonomously retrieve and synthesize evidence without task-specific training. Finally, to explore the performance ceiling under the rigorous exact match accuracy metric, we evaluate a trained version of EHRNavigator (trained). In this configuration, the answer-generation agent is optimized using 80% of the DrugEHRQA dataset. This tuning is specifically designed to enhance formatting consistency and stylistic alignment, ensuring that correctly retrieved evidence is translated into the precise standardized output required by the benchmark, while the underlying evidence extraction pipeline remains identical to the no training version.

Figure 2 (a) summarizes the accuracy achieved by each model under these settings on DrugEHRQA. In the Vanilla setting, LLaMA3-70B-Instruct performs poorly at 10.21%, while LLaMA3.1-70B-Instruct and GPT-4o achieve 36.86% and 39.93%, respectively. MultimodalQA, which is a pre-trained language model with ClinicalBERT as the backbone, yields a higher accuracy of 61.08%. When equipped with EHRNavigator (no training), the accuracy rises to 33.78% for LLaMA3-70B-Instruct, 47.02% for LLaMA3.1-70B-Instruct, and 51.04% for GPT-4o. The EHRNavigator (trained) variant further improves accuracy to 79.78% for LLaMA3-70B-Instruct and 81.32% for LLaMA3.1-70B-Instruct. Figure 2(b) shows the absolute improvements introduced by EHRNavigator over the Vanilla LLMs. LLaMA3-70B-Instruct gains 23.57 percentage points, GPT-4o improves by 11.11 points, and LLaMA3.1-70B-Instruct improves 10.16 points. After training the answering model in EHRNavigator, these gains are significantly amplified, LLaMA3-70B-Instruct even improves by 69.57 points and LLaMA3.1-70B-Instruct gains 44.46 points over their respective baselines. The substantial performance boost in the trained version demonstrates that the model successfully learned to extract and format standardized answers that align with the rigorous requirements of the data. Overall, these findings validate EHRNavigator's capability to accurately answer patient-level clinical questions by effectively interacting with both structured and unstructured EHR data.

### 2.3 Performance of structured data querying

We further evaluate the effectiveness of different approaches for generating accurate queries on structured data, specifically focusing on the execution accuracy of the resulting SQL on two datasets: DrugEHRQA [16] and EHRSQL [10]. We compare three distinct methodologies: (1) traditional text-to-SQL models, including TREQS [16] and T5 [10], which are trained on specific question-SQL pairs; (2) a non-agent Vanilla LLM prompting baseline where powerful LLMs are directly prompted with the full database schema; and (3) our proposed EHRNavigator multi-agent framework that decomposes structured query generation task into modular subtasks.

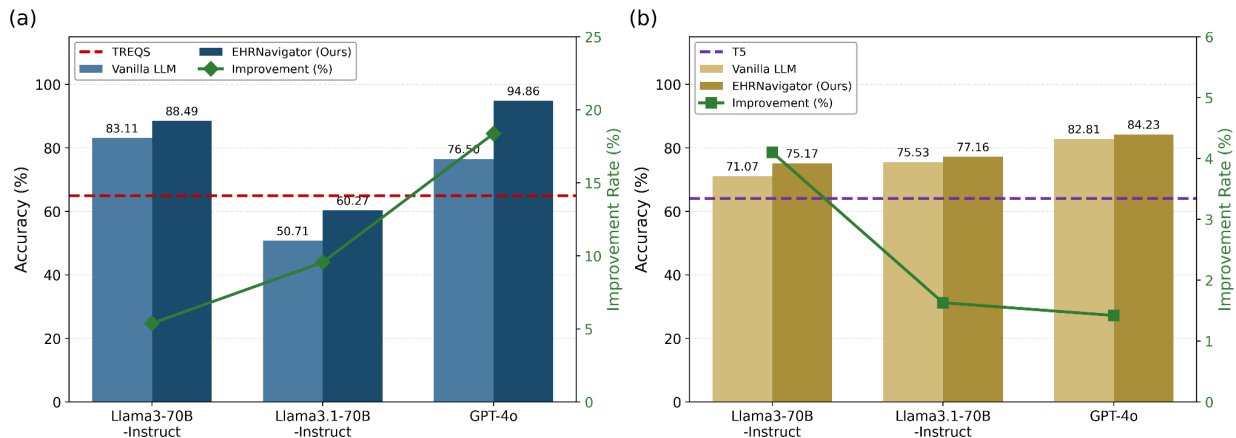

**Figure 3.** Performance analysis of EHRNavigator against vanilla LLM prompting and specialized Text-to-SQL baselines. (a): Execution accuracy on the DrugEHRQA dataset, comparing vanilla backbones and EHRNavigator against the TREQS [16] baseline (dashed line). (b): Execution accuracy on the EHRSQL dataset, comparing vanilla backbones and EHRNavigator against the T5 [10] baseline (dashed line). In both subfigures, bars (primary y-axis) indicate absolute accuracy (%), while markers and lines (secondary y-axis) represent the relative improvement of the percentage increase in accuracy achieved by the EHRNavigator framework over the corresponding vanilla LLM baseline.

The results of the evaluation on the structured data querying are summarized in Figure 3. Traditional text-to-SQL models established a foundational performance level, TREQS achieved 64.92% on DrugEHRQA, while T5 achieved 64.00% on EHRSQL after training on these datasets specifically. Compared with these results, the Vanilla LLM baseline using powerful LLMs generally yielded significantly higher initial accuracy without any further training. For instance, GPT-4o achieved 76.50% on DrugEHRQA and 82.81% on EHRSQL, demonstrating the strong inherent capability of large models. Notably, LLaMA3-70B-Instruct reached a high baseline of 83.11% on DrugEHRQA, while LLaMA3.1-70B-Instruct showed lower initial performance on DrugEHRQA at 50.71% yet achieved a more competitive 75.53% on EHRSQL. This discrepancy primarily stems from the model's tendency to omit medication units in the generated queries, which frequently led to exact match accuracy failures under the rigorous evaluation criteria of DrugEHRQA.

Subsequently, we assessed the performance of EHRNavigator. It consistently resulted in substantial performance gains across all LLM backbones compared to their respective Vanilla baselines. The most dramatic enhancement was observed on the DrugEHRQA dataset, where the use of GPT-4o saw its accuracy surge from 76.50% to 94.86%, marking a remarkable absolute improvement of 18.36%. Similarly, LLaMA3.1-70B-Instruct, which had the lowest baseline, realized an impressive 9.56% gain, increasing its accuracy to 60.27%. On the EHRSQL dataset, while baselines were already strong, EHRNavigator still provided consistent positive improvements, notably boosting LLaMA3-70B-Instruct by 4.10% (from 71.07% to 75.17%) and GPT-4o by 1.42% (reaching 84.23%). This consistent superiority validates the effectiveness of decomposing the complex structured query generation task into modular, manageable subtasks for LLMs.

### 2.4 Performance of unstructured data query

To verify the effectiveness of EHRNavigator in retrieving relevant unstructured information, we compared its performance against two baselines: Vanilla LLM (no-retrieval, where clinical notes are concatenated as raw context) and standard Retrieval-Augmented Generation (RAG) [23]. This evaluation utilized two distinct benchmarks: DrugEHRQA [16] and EHRNoteQA [23]. For DrugEHRQA, we focused on questions requiring entity extraction from clinical notes, measured via exact match accuracy. Conversely, EHRNoteQA involves questions requiring complex synthesis of free-text narratives (e.g., *"How did the patient feel upon admission?"*). Given the semantic flexibility of these answers, we employed generative evaluation metrics, ROUGE-L, BERTScore [38], and BARTScore [39], to capture reasoning quality beyond simple exact string matching.

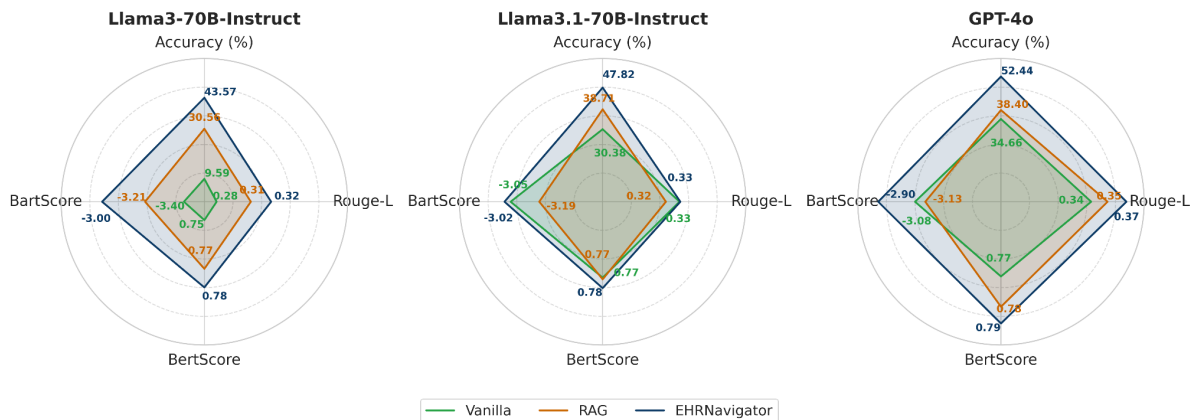

**Figure 4.** Comparative performance across different settings and LLM backbones. The figure presents a normalized score comparison of the Vanilla, RAG [23], and EHRNavigator approaches. Performance is evaluated across three distinct backbone models: LLaMA3-70B-Instruct, LLaMA3.1-70B-Instruct, and GPT-4o. The metrics displayed include Accuracy (on DrugEHRQA dataset) and the generative evaluation scores of ROUGE-L, BertScore, and BartScore (on EHRNoteQA dataset).

The results in Figure 4 highlight the essential nature of the retrieval mechanism for accurate question answering in real-world EHR environments. On the DrugEHRQA dataset, RAG methods significantly outperformed the no-retrieval baseline; for instance, GPT-4o's accuracy improved from 34.66% to 44.48%, while LLaMA3.1-70B-Instruct rose from 30.38% to 43.08%. The necessity of retrieval is most evident in models with restricted context windows, such as LLaMA3-70B-Instruct (8,192 tokens). In the Vanilla setting, the massive volume of clinical notes frequently exceeded this limit, leading to a low accuracy of 9.59%. However, with retrieval, its performance surged to 30.56%, underscoring the critical nature of context management. EHRNavigator further demonstrates a clear performance advantage. EHRNavigator further established a clear performance advantage, achieving 52.44% and 47.82% for GPT-4o and LLaMA3.1-70B-Instruct respectively. EHRNavigator also excels in textual analysis scenarios. On EHRNoteQA, EHRNavigator with GPT-4o achieved the best generative scores across the board: ROUGE-L (0.3664), BertScore (0.7910), and BartScore (-2.90). This superior performance confirms the robustness of its semantic retrieval component when tackling complex clinical narratives and generating nuanced answers.

**2.5 EHRNavigator returns essential evidence from both structured and unstructured EHR sources**

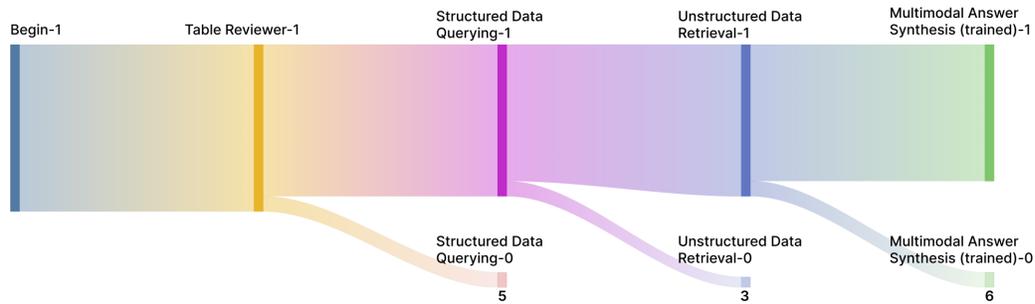

**Figure 5.** Error propagation and synthesis bottleneck analysis with manual review. It illustrates the flow of a set of questions through the EHRNavigator pipeline, tracking where errors accumulate (paths ending in 0) and where correct answers proceed (paths ending in 1). The data is derived from a manual review of 100 cases selected from the DrugEHRQA dataset. In this review, we assessed the output of every step of the pipeline to determine if the retrieved evidence (from structured querying or unstructured retrieval) contained the necessary information to derive the correct final answer or the final answer is correct or not.

To gain deeper insights into the performance bottlenecks of the EHRNavigator framework, we conducted a rigorous manual review of 100 cases selected from the DrugEHRQA dataset. As illustrated in Figure 5, this analysis tracks the flow of clinical queries through the pipeline, identifying where correct reasoning paths divert into errors (marked as 0) and where they proceed toward a successful outcome (marked as 1). By assessing the output of every modular step, we characterized the framework's ability to maintain data fidelity from initial retrieval to final synthesis.

The analysis reveals that EHRNavigator's underlying retrieval mechanisms are robust. Minimal error accumulation occurred during the initial evidence-gathering stages, with only 5 errors identified in Structured Data Querying and 3 errors in Unstructured Data Retrieval. This confirms that the framework's core strength lies in its ability to navigate complex EHR schemas and narratives to surface high-fidelity clinical evidence. The Multimodal Answer Synthesis stage further observed 6 error instances, where the models generated incorrect final answers despite being provided with the necessary evidence. This "synthesis gap" highlights that while the system is highly effective at gathering foundational data, the challenge of multimodal evidence reconciliation remains non-trivial for LLMs. Collectively, these results validate that EHRNavigator provides clinicians with a robust foundation for decision-making by ensuring the retrieval of comprehensive evidence, while also demonstrating that targeted training can effectively bridge the final synthesis gap to achieve superior end-to-end accuracy.

**2.6 Performance on real patients' trajectories on YNHHQA**

To assess the generalizability of EHRNavigator in a live clinical environment, we deployed the system on real-world data from YNHH. Unlike the MIMIC-III data used in previous experiments, the YNHH database follows the OMOP Common Data Model (CDM), presenting a significant architectural shift of the EHR structure. Specifically, we evaluated the system's ability to construct longitudinal patient trajectories, a task identified by clinicians as a primary bottleneck in manual chart review [48]. A benchmark of 100 physician-curated clinical questions was constructed and validated via chart review to verify the precision of the framework's outputs.

These questions were categorized into three groups: lab-related, drug-related, and lab–drug correlation questions. The corresponding results are summarized in Figure 5. On the 100-item physician-curated patient-trajectory benchmark, EHRNavigator achieved an overall accuracy of 86%. Accuracy by question type was 87.5% for laboratory trajectory questions, 80.0 % for medication dosing or timing questions, and 95.0% for combined lab–medication relations. Notably, the accuracy for combination queries surpassed that of single-domain tasks. This counter-intuitive finding is due to the clear structured clinical patterns inherent in such questions. While isolated medication histories often require navigating complex visit-level hierarchies, combination queries typically focus on observing physiological trends following a discrete intervention (e.g., drug administration). These explicit temporal and causal anchors constrain the search space, allowing the LLM to execute more precise queries.

A granular audit of the 14 identified failures revealed that these errors were primarily driven by the structural complexities of real-world EHR data rather than fundamental reasoning flaws. Encounter misalignment constituted most of these errors (57.1%), typically occurring when a query targeted a specific encounter that lacked the requested data. For instance, when tasked with analyzing laboratory trends during a patient's last visit, the system occasionally drifted into adjacent temporal windows if no relevant records existed within the specified terminal encounter. Such "null-data" scenarios often triggered the retrieval agent to search previous valid records to provide an answer, leading to a mismatch between the user's temporal constraint and the returned evidence. Information displacement within semi-structured fields accounted for 21.4% of the failures, where critical clinical facts were absent from standardized structured columns but embedded within a column with unstructured information (e.g., medication instruction), necessitating a shift to semantic text parsing that the initial SQL-based retrieval was constrained by structured schema limitations. Other failure modes included complex fragmented synthesis (7.1%), where the system struggled to bridge the gap between drug concentrations embedded as text in medication names (e.g., "20mg/ml") and volumes in structured quantity fields, and information overflow (7.1%), where a high density of historical records led to incomplete answer synthesis. At last, one case (7.1%) was identified as a pure reasoning hallucination, where the system misinterpreted the retrieved evidence despite offering the comprehensive evidence presented together with the answer.

Despite these initial omissions, a follow-up assessment of clinical recoverability demonstrated that most of these cases were resolvable through minimal clinician interaction. We identified two primary recovery pathways depending on the error type. First, by refining queries with clearer temporal or setting constraints, such as specifying 'inpatient visit' instead of a generic 'last visit', clinicians could effectively redirect the retrieval agent to the correct data window. Second, for complex synthesis or calculation errors, clinicians achieved success by decomposing inquiries into atomic, more general components. For instance, rather than requesting a calculated daily aspirin dosage, clinicians queried the medication name, individual drug quantity and time. This iterative probing allowed for the successful retrieval of the ground truth in all instances, suggesting that the system serves as a highly robust collaborative tool. Ultimately, these findings indicate that while EHR data fragmentation poses barriers to full automation, the system's interactiveness effectively mitigates these challenges in real-world clinical workflows.

We measured end-to-end latency across three query categories to evaluate system efficiency. As illustrated in Figure 6(b), the median latency for laboratory values was 11.53 s (IQR 9.91–16.38 s), which was comparable to the performance of drug dosing/timing queries at 15.47 s (IQR 11.78–17.94 s). Both single-domain query types exhibited similar upper-whisker limits, suggesting a consistent computational baseline for standard retrieval tasks. In contrast, the laboratory-drug combination queries were the most computationally intensive, with a median latency of 17.95 s. This is primarily driven by the mandatory execution of high-cardinality joins across visit, drug, and lab tables. The system prioritizes structural data alignment, synchronizing temporal medication logs with lab results, before initiating LLM reasoning. This deliberate pre-processing overhead ensures that the model operates on a filtered, high-fidelity context, effectively explaining the superior accuracy observed in Figure 6(a).

Furthermore, we assessed the operational expenditure for patient-trajectory queries. From the results in Figure 6(c), integrated lab–medication queries incurred the highest costs due to the high data density inherent in cross-event trajectories. Cross-referencing extensive, multi-row administration logs with dense longitudinal laboratory results significantly expands the LLM input context window. When both domains contain voluminous records, the simultaneous "JOIN" retrieval leads to a compounded boost in token consumption, necessitating more extensive reasoning cycles. This phenomenon reveals a fundamental trade-off: while deeper clinical synthesis through the integration of pharmacological and physiological information is desirable, it also indicates that future system designs should avoid high-cost JOIN operations to reduce unnecessary computational overhead and improve scalability in production environments.

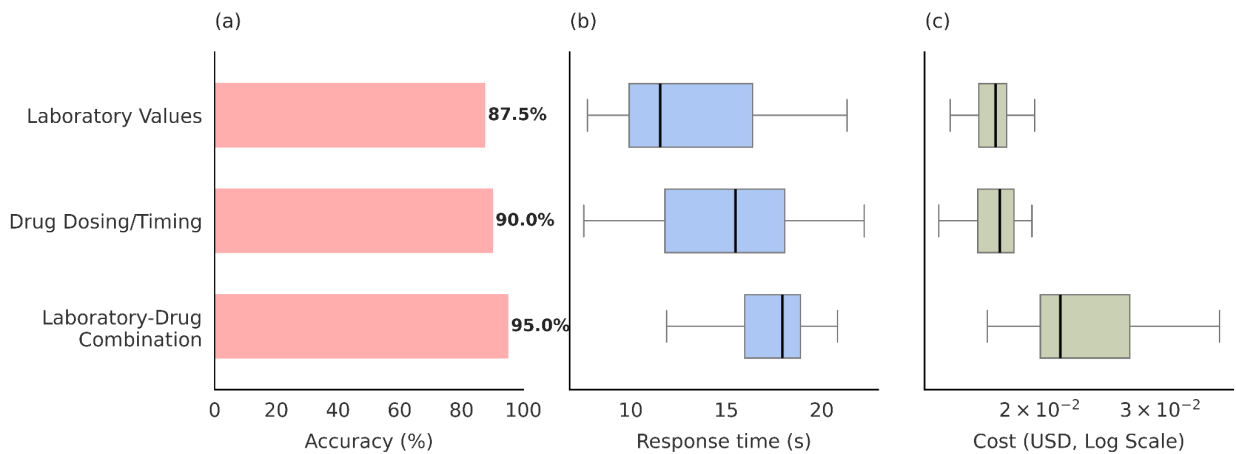

**Figure 6**. Accuracy and response time of EHRNavigator across question types on the Yale patient-trajectory benchmark (100 physician-curated questions). (a). Accuracy (%) for laboratory values, drug dosing/timing, and laboratory–drug Combination questions. (b). Distribution of end-to-end response time for the same categories on a linear scale, with black vertical lines indicating the median and blue boxes representing the interquartile range. (c) Cost distribution of end-to-end expenditure (USD) per question type, visualized on a logarithmic scale to highlight variances across orders of magnitude.

## 3. Discussions

In this study, we propose EHRNavigator, an end-to-end multi-agent framework for patient-level clinical question answering that seamlessly integrates structured and unstructured EHR data modalities. The technical innovation of EHRNavigator lies in its unified approach to integrating three essential yet typically isolated components: structured data querying through schema-aware SQL generation, unstructured data retrieval via chunking and semantic search, and answer synthesis through cross-modal evidence integration.

### 3.1 Flexibility of EHRNavigator

A major strength of EHRNavigator lies in its exceptional adaptability to diverse database schemas. Unlike traditional systems that require extensive manual configuration or schema-specific guidance [34, 9], EHRNavigator operates in a zero-shot manner across diverse database schemas by employing tool-augmented agents capable of directly interacting with databases and dynamically interpreting their structures at runtime. This design eliminates the need for schema-specific training or hard-coded mappings, allowing for rapid deployment across institutions and EHR platforms with minimal configuration. By generalizing effectively across heterogeneous data infrastructures, EHRNavigator provides a scalable foundation for cross-site clinical applications (from MIMIC to OMOP) and fosters interoperability in real-world healthcare settings.

### 3.2 Advances of EHRNavigator in multi-modal information integration

Another key advantage of EHRNavigator is its ability to holistically analyze patient information by seamlessly integrating both structured (e.g., lab results, medications, vitals) and unstructured (e.g., clinical notes, discharge summaries) data sources in near-real time. In contrast to traditional EHR systems such as Epic, where clinicians often rely on keyword or fuzzy searches and manual navigation through fragmented records, EHRNavigator leverages retrieval-augmented reasoning to automatically identify, aggregate, and synthesize relevant evidence in response to complex clinical questions. Beyond producing concise answers, the system also provides traceable supporting evidence from underlying records, enabling clinicians to verify conclusions and better understand the rationale behind each response. This multimodal, evidence-grounded reasoning process significantly enhances clinical efficiency and decision transparency.

### 3.3 Real-world deployment and practical implications

The evaluation on Yale New Haven Hospital data represents a critical validation of EHRNavigator's real-world applicability. Achieving 86% accuracy on physician-curated questions with a median response time of 12.16 seconds demonstrates practical feasibility for clinical workflows. Recent reviews have emphasized that evaluation methodologies for LLMs in healthcare must prioritize clinically relevant concerns over exact string matching [40]. Our physician-validated evaluation protocol addresses this by requiring detailed chart review rather than relying solely on automated metrics, ensuring that system outputs meet clinical standards for accuracy and completeness.

Crucially, our evaluation moves beyond traditional data-point queries, which focus on isolated facts to emphasize longitudinal patient trajectories. In existing benchmarks, tasks are often reduced to simple retrieval. However, in actual clinical practice, the primary bottleneck for physicians is the synthesis of a patient's history over time. Trajectory analysis is significantly more time-consuming for clinicians as it requires querying, filtering, and comparing multiple clinical streams (e.g., correlating medication titration with lab response) across disparate hospital visits. By automating this reconstruction, EHRNavigator addresses a high-value clinical need that traditional point-lookup systems ignore. The observed error patterns provide insights for future improvements. Visit-scope mismatches accounted for most of the errors, highlighting the challenge of temporal reasoning in complex multi-visit scenarios. These findings suggest that incorporating explicit temporal relation modeling and visit-context disambiguation could further enhance performance.

### 3.4 Limitations and future directions

EHRNavigator's current evaluation focuses primarily on fact-based queries tied to specific clinical entities (e.g., "*What was the patient's WBC count?*"), while many real-world clinical questions require higher-level reasoning about disease categories or treatment patterns (e.g., "*Does the patient have structural lung disease?*"). Extending EHRNavigator to handle concept-driven queries will require enhanced semantic understanding and potentially integration with medical ontologies like SNOMED-CT or UMLS. Additionally, the single-round querying paradigm may be insufficient for complex cases requiring iterative hypothesis refinement; a multi-turn dialogue capability would enable more sophisticated clinical reasoning akin to physician thought processes.

The generalizability demonstrated across MIMIC and OMOP schemas is encouraging, but further validation across additional EHR systems (e.g., Epic, Cerner) and clinical specialties would strengthen evidence for broad applicability. Recent evaluations of LLMs for health question answering have shown substantial variation across domains and question types [42], underscoring the need for comprehensive evaluation across diverse clinical contexts.

Beyond scaling with larger medical models to serve as backbones [41], our analysis reveals that existing LLMs still exhibit a fragmented understanding of complex temporal relationships between clinical events and patient visits. The inherent difficulty of aligning longitudinal data across multiple visits, evidenced by the visit-scope mismatch rate still remains a significant bottleneck. Future research should focus on enhancing the understanding of Temporal-Clinical Events. By explicitly modeling clinical chronologies, such a module would provide the framework with a robust temporal anchor, transforming EHRNavigator into a platform capable of interpreting complex patient trajectories with high precision.

## 4. Conclusions

EHRNavigator represents a significant step toward making EHR data more accessible and interpretable for clinical decision-making. By enabling natural language interaction with both structured tables and unstructured notes, the system has the potential to reduce cognitive burden on clinicians, decrease time spent on chart review, and minimize information retrieval errors, addressing critical challenges identified in EHR usability research. The framework's modular, agent-based architecture provides a flexible foundation that can readily incorporate new retrieval strategies, additional data modalities (e.g., medical images, time-series physiological data), or specialized reasoning modules. We anticipate that such multi-agent frameworks will become integral to next-generation clinical decision support systems, bridging the gap between the vast information contained in EHRs and the actionable insights needed at the point of care.

## 5. Materials and Methods

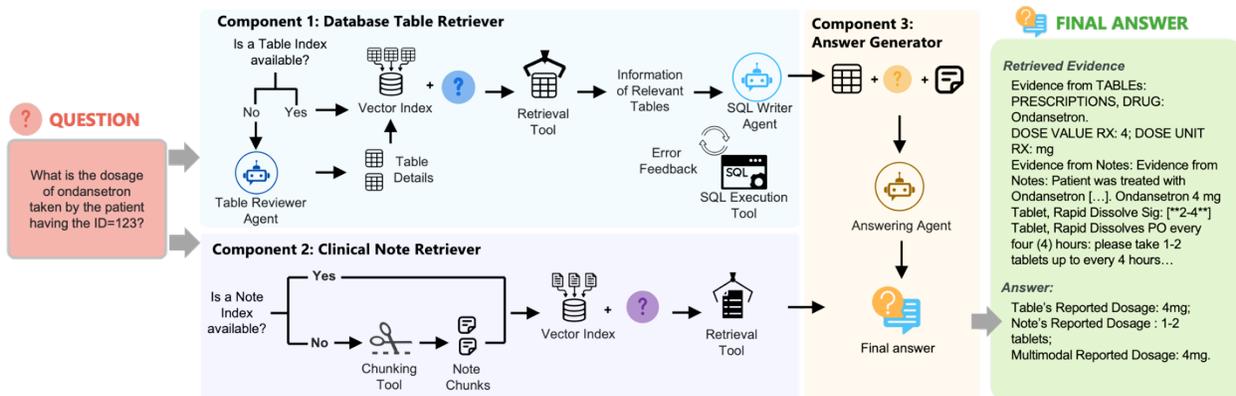

**Figure 6:** The overall framework of EHRNavigator. Given a clinical question, EHRNavigator activates a coordinated multi-agent system composed of specialized agents and external tools across three components. First, the structured data querying component uses a table reviewer agent, table retrieval tool, SQL generation agent and execution tool to explore the database schema, identify relevant tables, generate correct SQL query, and execute context-aware queries. Second, the unstructured data retrieval component leverages the structured query results and the original question to retrieval tools in locating relevant clinical notes. At last, the answer synthesis component employs a fine-tuned answer generator agent to integrate multimodal evidence and generate a coherent, clinically meaningful answer.

We introduce EHRNavigator, as illustrated in Figure 6, an agent-orchestrated, end-to-end multi-agent framework for patient-level question answering with EHR data. It consists of three core modules: (1) a structured data querying module that interprets natural language questions, understands database schemas, and generates SQL queries to extract information from structured tables; (2) an unstructured data retrieval module that identifies relevant content from clinical notes using chunking and semantic search; and (3) an answer synthesis module that integrates evidence from both modalities to produce accurate, context-aware, and explainable answers. By distributing responsibilities across multiple agents and dynamically leveraging tool capabilities, EHRNavigator supports natural language interaction without requiring pre-defined templates, manual query construction, or schema-specific customization, making it a flexible and scalable solution for real-world clinical environments.

### 5.1 Problem formulation

In this paper, we study the problem of patient-level EHR question answering. The input consists of a patient's complete EHR, including a collection of structured tables identified by their table names $T = \{t_1, t_2, \ldots, t_m\}$, where each table $t_i$ is associated with a set of columns $C_i = \{c_i^1, c_i^2, \ldots\}$, as well as a set of unstructured clinical notes $N = \{n_1, n_2, \ldots, n_k\}$. Given a clinical question $q$, the goal is to develop a question answering system $f$, that generates an answer by jointly reasoning over structured and unstructured EHR data. Finally, we formalize the task as:

$$A = f(q, T, N).$$

### 5.2 EHRNavigator framework

#### 5.2.1 Structured data query module

Given a natural language query $q$, EHRNavigator first extracts information from structured tables $T$ by leveraging different agent characters.

**Table Structure Discovery and Description.** This module automatically discovers and summarizes the structure of the underlying EHR database. The system first checks whether table descriptions have been previously encountered and cached. If not, it initializes a Table Reviewer

Agent $\mathcal{A}_\mathcal{R}$ to analyze the database. For each table name $t_i \in T$, the agent queries the database schema to obtain structural metadata associated with the table, including its column names $C_i = \{c_i^1, c_i^2, ...\}$ and relational information $k_i$, including primary and foreign keys. Based on the retrieved schema information, the Table Reviewer Agent generates a concise natural-language description for each table:

$$D_i = \mathcal{A}_\mathcal{R}(t_i, C_i, k_i \mid \text{prompt}_{\text{table-reviewer}}).$$

The resulting table descriptions $\{d_1, d_2, ..., d_m\}$ are cached for future reuse, enabling downstream agents, such as the SQL Writer Agent, to perform more efficient and concise query generation.

**Relevant table selection.** Next, EHRNavigator performs relevant table selection to identify tables that are most pertinent to a given clinical question. To support efficient retrieval, the system constructs semantic indexes over the table descriptions $\{d_1, d_2, ..., d_m\}$ generated in the previous stage. Given a question $q$, the system invokes a Table Retrieval Tool $\mathcal{T}_{ret}$ to match the question against the indexed table descriptions. The retrieval tool encodes both the question and each table description into a shared embedding space. Formally, let $\phi(\cdot)$ denote an embedding function. The semantic relevance score between the question and table $t_i$ is computed as

$$s_i = \text{sim}(\phi(q), \phi(d_i)),$$

where $\text{sim}(\cdot,\cdot)$ denotes a similarity measure such as cosine similarity. Based on these relevance scores, the table retrieval tool selects the top-ranked tables:

$$T_q = \mathcal{T}_{ret}(q, \{d_i\}_{i=1}^m) = TopK(\{t_i\}_{i=1}^m; \{s_i\}_{i=1}^m),$$

where $T_q \subseteq T$ denotes the set of tables most relevant to the question. The selected tables are then passed to downstream modules for SQL generation and structured reasoning.

**SQL generation.** Once the relevant tables are identified, the SQL generation module composes executable SQL queries based on the input question and the selected tables. Specifically, for each table name $t_i \in T_q$, a SQL Writer Agent $\mathcal{A}_\mathcal{W}$ is responsible for generating a query conditioned on the question and the table context. Beyond utilizing table names and column definitions, the agent further queries representative table contents to better understand the data semantics. For each selected table $t_i$, the agent invokes the SQL Sampler tool $\mathcal{T}_{SQL}$ to retrieve a sample row and column names:

$$\mathcal{T}_{SQL}(t_i) = (C_i, R_i),$$

where $C_i$ denotes the column list of tables $t_i$, and $R_i$ denotes a representative sampled row. Conditioned on the question and the selected tables $T_q$, the SQL Writer Agent generates executable SQL queries by reasoning over the table names, column information, and sampled rows associated with the selected tables. Formally, for each $t_i \in T_q$, we have

$$SQL_q = \mathcal{A}_\mathcal{W}\left(q, \{t_j, C_j, R_j\}_{t_j \in T_q} \mid \text{prompt}_{\text{sql-writer}}\right).$$

**Output generation.** The generated SQL query $SQL_q$ is subsequently executed against the database to retrieve structured results. Specifically, the system invokes a SQL execution tool $\mathcal{T}_{exec}$ to run the query:

$$E_q^s = \mathcal{T}_{exec}(SQL),$$

where $E_q^s$ denotes the evidence from the structured tables, which is the execution result returned by the database. If the execution fails due to syntax errors, schema mismatches, or missing data, the SQL Writer Agent would be required to generate another round of SQL. This iterative execution loop allows the system to progressively correct invalid queries and improves the robustness of SQL generation in the presence of complex schemas and noisy data.

#### 5.1.2 Unstructured data retrieval module

To retrieve relevant information from unstructured clinical notes, EHRNavigator employs two complementary tools coordinated by an agent: a Note Indexer for chunking and indexing, and a Note Retriever for structure-aware semantic search conditioned on the clinical question and structured evidence obtained from the Structured Data Module.

**Chunking and indexing (Notes indexer).** For a given patient, the system first checks whether a chunked index over clinical notes already exists.

If not, the agent invokes a note chunking tool $\mathcal{T}_{ch}$ to preprocess the notes. Specifically, given a set of clinical notes $N = \{n_1, n_2, \ldots, n_k\}$, the indexer segments each note into a set of smaller, semantically coherent text units:

$$\mathcal{U} = \mathcal{T}_{ch}(N),$$

where $\mathcal{U} = \{u_1, u_2, \ldots, u_L\}$ denotes the resulting collection of note chunks. Chunking is performed based on token length constraints and sentence boundaries, ensuring that each chunk remains contextually informative. Then, the chunks are indexed for further retrieval. This on-demand indexing strategy avoids precomputing note chunks and enables scalable retrieval in large-scale EHR systems, with preprocessing automatically triggered by the agent when needed.

**Structure-guided note retrieval (Notes retriever).** After indexing, EHRNavigator employs a structure-aware note retrieval tool $\mathcal{T}_{ret}$ to identify note chunks relevant to the input question. Unlike standard semantic retrieval, this tool conditions retrieval not only on the question $q$, but also on structured evidence $E_q^s$ obtained from SQL execution. The note retriever ranks note chunks by jointly encoding the question, structured evidence, and note chunks:

$$s_j = \text{sim}\left(\phi(q, E_q^s), \phi(u_j)\right), \quad u_j \in \mathcal{U},$$

Based on these relevance scores, the retrieval tool selects the top-ranked note chunks:

$$E_q^u = \mathcal{T}_{ret}(q, E_q^s, \mathcal{U}) = TopK\left(\mathcal{U}; \{s_j\}_{j=1}^L\right),$$

where $E_q^u \subseteq \mathcal{U}$ denotes the set of retrieved note chunks.

By incorporating structured query results into the retrieval context, this structure-guided retrieval strategy narrows the search space and improves alignment between clinical questions and unstructured documentation. Compared to conventional semantic retrieval methods, it more effectively bridges the semantic gap between user queries and clinical narratives, thereby supporting more accurate and reliable downstream answer synthesis.

#### 5.1.3 Answer synthesis

After collecting evidence from both structured tables and unstructured clinical notes, EHRNavigator invokes an Answer Synthesizer Agent $\mathcal{A}_{ans}$ to produce a single, clinically grounded response. The agent operates over a unified evidence context that integrates information from multiple modalities, including: (i) the executed SQL results returned by the SQL Execution Tool, and (ii) the top-K retrieved note chunks obtained from the unstructured data retrieval module. Formally, with the evidence from tables $E_q^s$ returned by SQL execution, which includes field names, values, and associated timestamps, and a set of retrieved note chunks with their corresponding timestamps $E_q^u$. The Answer Synthesizer Agent generates the final answer by conditioning on the joint evidence:

$$A = \mathcal{A}_{ans}(q, E_q^s, E_q^u \mid prompt_{answer-synthesizer}).$$

The underlying large language model integrates evidence across modalities to generate a coherent response. However, the retrieved inputs may contain redundancy, nested structures, or conflicting information, which poses challenges for direct generation and can lead to inconsistent or unreliable outputs [20, 21]. To enhance cross-modal reasoning and conflict resolution, we fine-tune the underlying language model using expert-curated data from DrugEHRQA [16]. During this stage, paired structured and unstructured evidence $(E_q^s, E_q^u)$ together with the question q are serialized into a single input context, and the agent $\mathcal{A}_{ans}$ is trained to generate the integrated ground-truth answers.

The fine-tuning follows an instruction-following paradigm, explicitly guiding the model to produce concise, clinically grounded summaries that reconcile potentially conflicting evidence. We apply supervised fine-tuning with a cross-entropy objective and adopt low-rank adaptation (LoRA) modules [33] to ensure computational efficiency. Importantly, this training procedure is applied only to the answer synthesis agent $\mathcal{A}_{ans}$. All upstream components, including the structured data querying module and the unstructured data retrieval module, remain entirely unsupervised. This design choice isolates the evaluation of retrieval quality and demonstrates the generalizability of the retrieval pipeline without task-specific tuning. As a result, the agent $\mathcal{A}_{ans}$ exhibits improved ability to integrate heterogeneous clinical inputs, reason across modalities, and assess the reliability of evidence in the presence of inconsistencies, yielding responses that are more coherent, context-aware, and clinically interpretable in real-world patient scenarios.

### 5.2 Evaluation datasets

To evaluate EHRNavigator for patient-level question answering across different data modalities, we conducted experiments on three public MIMIC-derived datasets, DrugEHRQA (MIMIC-III), EHRSQL (MIMIC-III), and EHRNoteQA (MIMIC-IV), all based on de-identified EHRs from Beth Israel Deaconess Medical Center [2,22]. In addition, we included YNHHQA, a physician-curated benchmark comprising 100 patient-trajectory questions from Yale New Haven Hospital. Collectively, these datasets encompass structured, unstructured, and multimodal queries.

#### 5.2.1 DrugEHRQA (Multimodal)

DrugEHRQA is a large-scale multimodal question answering dataset built on MIMIC-III, focusing on drug-related queries by leveraging both structured EHR tables and unstructured clinical notes [16]. Each question is annotated with SQL queries/answers from structured data, unstructured notes, and a merged multimodal answer, supporting robust evaluation across modalities. To ensure consistency with prior work, we adopt the same experimental setup and use the subset specifically designed for multimodal evaluation [16]. This subset includes 5,559 structured QA examples that can only be answered using structured tables, 5,693 unstructured QA pairs with answers sourced from clinical notes, and 27,795 multimodal question–answer pairs that require both structured and unstructured data to obtain accurate answers.

### 5.2.2 EHRSQL (Structure)

It is a text-to-SQL benchmark built on the MIMIC-III database, featuring diverse and realistic clinical questions grounded in real-world scenarios [10]. The dataset emphasizes temporal reasoning, complex SQL structures, and supports the development and evaluation of robust medical text-to-SQL models. Since the original dataset does not include gold execution results to protect patient privacy. We first execute the gold SQL queries to obtain the corresponding answers. In this step, we filter those questions where the gold SQL failed to execute over the database because execution time exceeds. This yields a set of 1,410 question–answer pairs with executed results, which we use for evaluation to ensure consistency and correctness.

### 5.2.3 EHRNoteQA (Unstructured)

EHRNoteQA is a benchmark for clinical question answering over discharge summaries, based on the MIMIC-IV dataset [23]. It consists of 962 high-quality QA pairs spanning 10 clinically meaningful categories, such as treatment, assessment, and etiology. Each question requires reasoning over one or more discharge summaries for a given patient, capturing the complexity of real-world longitudinal clinical scenarios. The dataset focuses exclusively on unstructured clinical notes, offering a challenging and practical evaluation setting for models designed to reason over free-text EHR narratives. Since real-word clinical questions are open-ended, we convert each multiple-choice item into a free-text pair by extracting the correct option as the reference answer, yielding an open-ended evaluation format while preserving the original supervision.

### 5.2.4 YNHHQA

To further validate longitudinal reasoning in a real-world setting, we evaluated the system on EHR data from Yale New Haven Hospital. While the 100 clinician-curated patient-trajectory questions were originally authored over MIMIC encounters, this evaluation was conducted using real YNHH EHR data organized under the OMOP common data model. This setup allows assessment of the model's robustness and generalizability beyond benchmark datasets and into production-grade clinical data environments.

The YNHHQA spans three question categories, laboratory trajectories, medication dosing and timing, and mixed temporal relations (e.g., drug–lab coupling or single-trajectories checks within specific encounters or time frames). This tripartite design draws upon prior work on temporal reasoning in clinical data and aims to capture the dynamic, time-dependent nature of real-world clinical decision-making, questions that clinicians frequently need to answer but often find difficult to obtain from existing EHR interfaces [36, 43]. Each question was executed against OMOP-formatted structured tables, with retrieved clinical notes incorporated when narrative reasoning was required. Representative examples include "*How did troponin change over the last ED visit?*" and "*Did vancomycin ever exceed 2 g/day?*".

For each question–case pair, our system generated an answer, which was then validated against the patient chart using a standardized chart review protocol. Specifically, physician reviewers determined answer correctness by verifying whether the response was supported by evidence in the EHR (structured data and/or clinical notes) within the pre-specified temporal window. Final correctness labels were assigned based on chart review. This evaluation represents a test of EHRNavigator's ability to operate over real data and demonstrates its capacity to produce accurate, context-aware, and clinically grounded answers in authentic clinical environments.

### 5.3 Experiment settings

**Backbones and agent roles.** We evaluate multiple large language models as backbones for different agent roles within EHRNavigator, including the closed-source GPT-4o (Azure-2024-10-21, 128k context) [24] and two open-source alternatives: Llama 3-70B-Instruct (8k context) and Llama 3.1-70B-Instruct (128k context) [25]. Each role, including Table Reviewer, SQL Writer, and Answer Synthesizer, uses a corresponding LLM to perform specialized subtasks, such as table description, SQL generation, and multimodal answer synthesis.

**Prompts and decoding settings**. For consistency, all models employ the same instruction templates (available in the Supplementary Materials). To ensure task alignment, each task uses a slightly adapted prompt tailored to its context for different datasets. The decoding temperature is fixed at 0 across all experiments to guarantee deterministic and reproducible outputs.

**Retrieval configuration.** Structured and unstructured retrieval pipelines share a unified retrieval backbone. For structured tables, we use the BGE-large-en[1] embedding model to encode both questions and table descriptions, selecting the top k as 10 for most relevant tables [26]. For unstructured clinical notes, we also leverage BGE-based semantic retrieval. Notes are segmented into 256-token chunks with a 32-token overlap using LlamaIndex's sentence-aware chunker[2], ensuring coherent boundaries. Given that most clinical questions are temporally grounded, each note chunk is concatenated with its corresponding timestamp to preserve temporal context. When structured evidence is unavailable, the system automatically falls back to note-based retrieval, applying temporal and section filters to refine the search.

**SQL execution and schema exploration.** SQL generation and execution are decoupled. Generated queries are executed against SQLite databases formatted in OMOP and MIMIC structure, with a 120-second timeout and strict parameter binding to prevent injection.

**Training setups.** To establish the performance upper bound of the framework, we fine-tuned the Answer Synthesizer using LoRA. We employed a rank of 16 and alpha of 32, targeting the query and value projection matrices within the transformer blocks. The model was trained using supervised fine-tuning with a learning rate of 1e-5 and a batch size of 8 for 1 epoch. The training data utilized 80% of the DrugEHRQA multimodal split. Upstream components, including the Table Reviewer and SQL Writer agents, remained entirely frozen in a zero-shot state to ensure the generalizability of the retrieval pipeline.

**Evaluation protocol and metrics.** For DrugEHRQA and EHRSQL, all questions are entity-focused or directly from executed results, we adopt accuracy as the evaluation metric. For each answer, if the model output exactly matches the ground truth, we consider it as correct. In contrast, EHRNoteQA consists of free-text answers that are often not tied to specific entities. To assess the semantic alignment between generated and reference answers in this dataset, we employ Rouge scores, BERTScore, and BARTScore as evaluation metrics. For YNHHQA, all results are first generated by EHRNavigator, and human reviewers then perform chart review on the corresponding patient records to determine the correctness of each answer.

**Baselines and ablation study design.** Our baselines and ablation settings are matched to the three major tasks evaluated in this study, multimodal QA, structured data query, and unstructured note-based QA. For multimodal QA, we compare different backbone models under three configurations: a Vanilla setting where models directly receive the discharge summary and relevant structured

---

[1] https://huggingface.co/BAAI/bge-large-en/
[2] https://www.llamaindex.ai/

tables; an EHRNavigator (no training) setting where the same models operate inside our multi-agent retrieval and decomposition pipeline; and an EHRNavigator (trained) setting where the answer-generation agent is further fine-tuned on DrugEHRQA. For structured SQL generation, we compare classical text-to-SQL models (TREQS and T5), direct schema-prompting of LLMs (Vanilla), and our decomposed EHRNavigator SQL pipeline. For unstructured QA, we contrast Vanilla where the entire clinical notes are given directly to the LLMs, a RAG framework where retrieval models are leveraged to filter out related chunks from clinical notes. These baselines collectively allow us to isolate the contributions of retrieval, decomposition, and targeted fine-tuning across all components of the system.

**Computation settings.** All experiments were executed on 2 × H100 (80 GB) GPUs. GPT-4o was accessed through the hospital-managed Azure OpenAI environment (API version 2024-10-21) for both MIMIC and YNHHQA. For YNHH evaluation, all patient data handling was performed under IRB approval within a secure, access-controlled platform (e.g., restricted network, role-based access control, and audit logging) to ensure rigorous protection of patient privacy and clinical data security.

## Data Availability

The DrugEHRQA dataset is publicly accessible via https://physionet.org/content/drugehrqa/1.0.0/. The EHRSQL dataset can be obtained from https://github.com/glee4810/EHRSQL. The EHRNoteQA dataset is available on https://physionet.org/content/ehr-notes-qa-llms/1.0.1/.

## Code Availability

The code will be available after the manuscript is accepted.

## Acknowledgment

## Supplementary

### Prompts used in our work

To accommodate the heterogeneous nature of EHR databases (e.g., the relational structure of MIMIC-III vs. the standardized OMOP CDM in YNHH), we employed task-adaptive instruction prompting. While the core reasoning logic of the SQL Writer and Answer Synthesizer agents remained consistent, the specific instructions (such as temporal unit handling or concept-id resolution) were tailored to the target dataset's schema requirements. This design ensures that EHRNavigator can navigate diverse institutional data environments without the need for manual database re-indexing or model fine-tuning.

Prompts for table descriptions generation

```
prompt = """Please provide a clear and concise description of the database table based on its structure:

Table Name: {table_name}

Columns:
{columns}

Primary Keys: {primary_keys}
Foreign Keys: {foreign_keys}

Generate a comprehensive description that explains:
1. The purpose of this table
2. The meaning of key fields

Description:
"""
```

Prompts for EHRSQL (structured data querying)

```
prompt = """Given an input question, generate a syntactically correct SQLite query that directly answers it using
only the tables and columns defined in the provided schema. The query must include all necessary calculations,
conditions, and logical operations required to produce a complete and correct result.
Never query for all columns from a specific table; only select the columns that are relevant to the question.
Pay close attention to use only the column names visible in the schema description.
Be careful not to query columns that do not exist.
Ensure each column is used from the correct table, and qualify column names with their table name when
necessary to avoid ambiguity.

Instructions:
1. Time-based reasoning:
   - Perform all time-related calculations directly in SQL.
   - Use days as the unit of time rather than hours.
   - When identifying the first or last event, sort by the start time column (not the end time).
2. Boolean existence checks:
   - For yes/no questions return a Boolean-like result.
3. Schema adherence:
   - Use only column names and tables present in the schema.
   - Qualify columns with their table name if needed to avoid ambiguity.
4. Column semantics:
   - Differentiate correctly between admission location and admission type based on the question.
5. Ordering for interpretability:
   - If ordering improves clarity, apply ORDER BY on the most relevant column (typically time or numeric values).
6. Output format:
   - Only output the final SQL query.
   - Do not include any explanations, comments, or additional text.

Following this format as output:
Question: Question here
SQLQuery: SQL query to run
SQLResult: Result of the SQLQuery
Answer: Final answer here

Only use tables listed below.
{schema}

Question: {query_str}
SQLQuery:
"""
```

Prompts for DrugEHRQA (structured data querying)

```
prompt ="""Given an input question, first create a syntactically correct {dialect} query to run, then look at the results
of the query and return the answer.
You can order the results by a relevant column to return the most informative or representative examples in the
database.
Never query for all columns from a specific table; only select the columns that are relevant to the question.
Pay close attention to use only the column names visible in the schema description.
Be careful not to query columns that do not exist.
Ensure each column is used from the correct table, and qualify column names with their table name when
necessary to avoid ambiguity.
Please include the full entity name and hadm_id in the SQL if there is any in the question.

Following this format as output:
Question: Question here
SQLQuery: SQL query to run
SQLResult: Result of the SQLQuery
Answer: Final answer here
```

```
Only use tables listed below.
{schema}

Question: {query_str}
SQLQuery:
"""
```

Prompts for YNHHQA

```
prompt = """Given an input question, generate a syntactically correct {dialect} SQL query that directly answers it.
Then, execute the query and interpret its results to provide the final answer.
If applicable, order the results by a relevant column (e.g., time or numeric value) to return the most informative
examples in the database.

Please include all relevant tables that could potentially provide useful information.
Carefully identify the key entity mentioned in the question and accurately select the table where it would reside.
Use only the column names present in the schema description—do not reference columns that do not exist.
Always ensure that columns are correctly associated with their corresponding tables and qualify them with table
names when needed to avoid ambiguity.
The SQL query must be generated within a single line following the exact format below.

Follow these additional instructions:
1. Evaluate whether the entity mentioned in the question exists in the EHR data. If uncertain, query all values from
the relevant column.
2. Review all tables carefully and include every table that may be relevant to the question.
3. When searching for entity names (e.g., drugs, conditions), use the source value columns and apply `LIKE`
instead of `=` for matching. Use lowercase for comparison.
4. Expand each key entity (drug name, condition, hospital stay type, etc.) with all possible variations (abbreviations,
short forms, full names) using `OR` conditions for robust search.
5. Use `JOIN` only when necessary to resolve concept names from concept ID tables—not to combine unrelated
tables such as drugs and labs.
6. Always include the relevant time column in your query, as temporal information is crucial for clinical decision
support.
7. Include lab, drug, or other relevant tables that could help answer the question, when appropriate.
8. Do not use the NOTE table in any query.
9. For visit types, resolve the visit concept ID via the concept table and use the concept name to identify specific
visit types.
10. Return the output strictly in JSON format, with the following structure:
   {
      "SQL": "Generated SQL query"
   }

Following this format as output:
Question: Question here
SQLQuery: SQL query to run
SQLResult: Result of the SQLQuery
Answer: Final answer here

Only use tables listed below.
{schema}

Question: {query_str}
SQLQuery:
"""
```

Prompts for answer synthesize for DrugEHRQA:

```
prompt = """Given an input question, answer the question based on the query results.
```

You should also output the given SQL query and notes information. Only answer with the returned information.
Please pay attention to the time thread and rethink the time influence when you answer the question.
If the time in notes does not match time in the question, ignore them.
Do not make up something.
Please provide as much detail as possible when you answer the question.

Input:
- Query: {query_str}
- SQL Query: {sql_query}
- SQL Response: {context_str}
- Notes: {notes}

Output:
1. SQL QUERY: Provide the SQL query used.
2. Evidence from notes: Provide some information from notes.
2. Response: Provide a concise and accurate response to the question based on the SQL response and notes, please include all the necessary details from the evidence to support your answer for the question.
"""

Prompts for EHRNoteQA:

```
prompt = """You are a highly knowledgeable assistant specializing in answering medical questions.
Please answer the question based on the context in one sentence.
CONTEXT: {enhanced_bge}
Question: {query}
OUTPUT:
"""
```